\documentclass[letterpaper]{article} 
\usepackage[]{aaai24}  
\usepackage{times}  
\usepackage{helvet}  
\usepackage{courier}  
\usepackage[hyphens]{url}  
\usepackage{graphicx} 
\usepackage{natbib}  
\usepackage{caption} 
\usepackage{algorithm}
\usepackage{algorithmic}

\usepackage{bm}
\usepackage{amsmath}
\usepackage{booktabs}
\usepackage{amssymb}

\usepackage{newfloat}
\usepackage{listings}
\DeclareCaptionStyle{ruled}{labelfont=normalfont,labelsep=colon,strut=off} 
\floatstyle{ruled}
\newfloat{listing}{tb}{lst}{}
\floatname{listing}{Listing}
\nocopyright 

\title{Biologically-Plausible Topology Improved Spiking Actor Network \\for Efficient Deep Reinforcement Learning}

\author{
    Duzhen Zhang\textsuperscript{\rm 1,2}\equalcontrib,
    Qingyu Wang\textsuperscript{\rm 1,2}\equalcontrib,
    Tielin Zhang\textsuperscript{\rm 1,2}\footnote{Corresponding author.},
    Bo Xu\textsuperscript{\rm 1,2,3}\footnotemark[2]
}
\affiliations{
    \textsuperscript{\rm 1}School of Artificial Intelligence, University of Chinese Academy of Sciences, Beijing, China\\
    \textsuperscript{\rm 2}Research Center for Brain-inspired Intelligence, Institute of Automation, Chinese Academy of Sciences, Beijing, China\\
    \textsuperscript{\rm 3}Center for Excellence in Brain Science and Intelligence Technology, Chinese Academy of Sciences, Shanghai, China\\

    $\{$zhangduzhen2019,wangqingyu2022,tielin.zhang,xubo$\}$@ia.ac.cn
}

\usepackage{bibentry}

\begin{document}

\maketitle

\begin{abstract}
The success of Deep Reinforcement Learning (DRL) is largely attributed to utilizing Artificial Neural Networks (ANNs) as function approximators. Recent advances in neuroscience have unveiled that the human brain achieves efficient reward-based learning, at least by integrating spiking neurons with spatial-temporal dynamics and network topologies with biologically-plausible connectivity patterns. This integration process allows spiking neurons to efficiently combine information across and within layers via nonlinear dendritic trees and lateral interactions. The fusion of these two topologies enhances the network's information-processing ability, crucial for grasping intricate perceptions and guiding decision-making procedures. 
However, ANNs and brain networks differ significantly. ANNs lack intricate dynamical neurons and only feature inter-layer connections, typically achieved by direct linear summation, without intra-layer connections. This limitation leads to constrained network expressivity.
To address this, we propose a novel alternative for function approximator, the Biologically-Plausible Topology improved Spiking Actor Network (BPT-SAN), tailored for efficient decision-making in DRL. 
The BPT-SAN incorporates spiking neurons with intricate spatial-temporal dynamics and introduces intra-layer connections, enhancing spatial-temporal state representation and facilitating more precise biological simulations.
Diverging from the conventional direct linear weighted sum, the BPT-SAN models the local nonlinearities of dendritic trees within the inter-layer connections. For the intra-layer connections, the BPT-SAN introduces lateral interactions between adjacent neurons, integrating them into the membrane potential formula to ensure accurate spike firing.
Comprehensive results from four continuous control tasks demonstrate that the BPT-SAN achieves superior network expressivity, surpassing both its artificial actor network counterpart and the regular SAN.
\end{abstract}

\section{Introduction}

Reinforcement Learning (RL) constitutes a machine learning paradigm that enables agents to learn and optimize action policies gradually through interactions with the environment, aiming to maximize rewards to the fullest extent possible~\cite{kaelbling1996reinforcement,sutton2018reinforcement}.
Nonetheless, conventional RL algorithms encounter obstacles when confronted with complex, high-dimensional state spaces.
To surmount this challenge, modern Deep Reinforcement Learning (DRL) leverages Artificial Neural Networks (ANNs) as robust function approximators, enabling the direct mapping of raw state space to the action space.
Consequently, DRL has yielded remarkable achievements across a spectrum of domains, encompassing financial trading~\cite{jiang2017deep}, gaming~\cite{mnih2015human,vinyals2019grandmaster}, robot control~\cite{duan2016benchmarking,lillicrap2016continuous,zhao2023ode}, and beyond.

A line of progresses in neuroscience has unveiled that efficient reward-based learning in the human brain involves the integration of spiking neurons with intricate spatial-temporal dynamics and network topologies featuring biologically-plausible connectivity patterns~\cite{schultz1998predictive,maass1997networks,douglas2004neuronal,stuart2016dendrites,ratliff1966dynamics}. 
This integration process allows spiking neurons to adeptly amalgamate information across various layers and within the same layer through dendritic trees and lateral interactions, respectively~\cite{wu2018improved,cheng2020lisnn}. 
The synergistic fusion of these two topologies empowers neuronal networks with heightened information processing capabilities, which is essential for engendering all-encompassing cognitive representations and facilitating decision-making~\cite{koch1983nonlinear}. 
Nevertheless, ANNs significantly diverge from brain networks. ANNs lack internal neuronal dynamics and only possess inter-layer connections, devoid of intra-layer connections.
Furthermore, in the inter-layer connections, ANNs commonly depict dendrite trees as basic linear structures, directly funneling weighted synaptic inputs to cell bodies. In contrast, physiological experiments reveal that actual dendritic branches harbor numerous active ion channels, contributing to intricate nonlinear characteristics that surpass the simplicity of a direct weighted sum~\cite{losonczy2008compartmentalized}.
All of these factors collectively impose limitations on the expressivity of ANNs.

In this paper, we introduce an innovative alternative for policy function approximator, known as the Biologically-Plausible Topology improved Spiking Actor Network (BPT-SAN), designed for efficient decision-making in DRL.
Unlike the conventional Artificial Actor Network (AAN), the BPT-SAN incorporates spiking neurons exhibiting intricate spatial-temporal dynamics and introduces intra-layer connections. 
This innovation empowers a more powerful spatial-temporal representation of states and enables a more precise simulation of biological computation.
Departing from the straightforward linear weighted sum characteristic of the typical AAN, the BPT-SAN models the local nonlinearity of dendritic trees in the inter-layer connections.
Illustrated in Figure~\ref{fig-BPT-SAN}, each post-synaptic neuron features multiple dendritic branches, which partition synaptic inputs in a random and exclusive manner.
These inputs are initially subjected to a weighted summation at each branch, followed by nonlinear integration to shape the final output for the post-synaptic neuron.
Moreover, in the intra-layer connections, the BPT-SAN incorporate lateral interactions between adjacent neurons, integrating them into the membrane potential formula to achieve more precise spike firing.

To ensure efficient learning, we train the BPT-SAN in conjunction with artificial critic networks, employing two cutting-edge policy-based DRL algorithms, TD3~\cite{fujimoto2018addressing} and SAC~\cite{haarnoja2018soft}, respectively, within the recently introduced hybrid learning framework~\cite{tang2020reinforcement,tang2021deep}.
Following the training phase, we comprehensively compare the BPT-SAN with its AAN counterpart and regular SAN (without specific network topologies) across four continuous control tasks sourced from OpenAI Gym~\cite{brockman2016openai}, including Hopper-v3, Walker2d-v3, Half-Cheetah-v3, and Ant-v3. 
We consider this to be a noteworthy endeavor aimed at enhancing SANs from a network topology perspective, moving towards more efficient decision-making, akin to the mechanisms observed in human brains.

Our primary contributions can be summarized as follows:
\begin{itemize}

\item Inspired by the brain networks, we propose a BPT-SAN for efficient decision-making in DRL, seamlessly integrating spiking neurons with intricate spatial-temporal dynamics and network topologies featuring biologically-plausible connectivity patterns.

\item The BPT-SAN models the local nonlinearity of the dendritic trees in the inter-layer connections and introduces lateral interactions between adjacent neurons in the intra-layer connections, significantly augmenting the information processing capability of the network.

\item Under consistent experimental configurations, the BPT-SAN, equipped with inter-layer localized nonlinearities and intra-layer lateral interactions, outperforms its AAN counterpart and regular SAN in each robot control task.

\end{itemize}

\section{Related Work}

\subsection{DRL}
Leveraging ANNs as function approximators, DRL adeptly handles complex, high-dimensional state spaces and ushers in rapid development~\cite{mnih2015human,silver2016mastering,vinyals2019grandmaster}.
Within DRL, two primary algorithmic families emerge: value-based~\cite{watkins1992q} and policy-based~\cite{sehnke2010parameter} approaches.

An exemplary representative of the value-based DRL algorithms is the Deep Q-Network (DQN)~\cite{mnih2015human}.
DQN employs ANNs to approximate the optimal Q function, leading to a strategy that selects actions maximizing the Q-value for a state.
Currently, DQN and its enhanced iterations, such as Dueling DQN~\cite{wang2016dueling} and Double DQN~\cite{van2016deep}, are making significant strides towards attaining human-level performance in Atari video games~\cite{bellemare2013arcade} featuring discrete action spaces. Another category of policy-based DRL algorithms excels in handling continuous action spaces.
The classic policy gradient algorithm is designed to directly learn the parameters within an artificial policy network~\cite{sehnke2010parameter}.
To ensure more stable learning, an advanced actor-critic framework is introduced. Here, an AAN deduces actions from given environmental states to represent the policy, while an artificial critic network estimates associated state values or action values to guide the actor in optimizing a better policy.
For further enhanced exploration and data efficiency, numerous actor-critic-based policy gradient algorithms have emerged, including A3C~\cite{mnih2016asynchronous}, TRPO~\cite{schulman2015trust}, PPO~\cite{schulman2017proximal}, DDPG~\cite{lillicrap2016continuous}, TD3~\cite{fujimoto2018addressing}, and SAC~\cite{haarnoja2018soft}.
Among these, TD3 and SAC stand out as the leading algorithms for robot control with continuous action spaces.

However, ANNs mimic brain networks roughly, conveying information primarily by firing rates and lacking the dynamic mechanisms within neurons.
In contrast, Spiking Neural Networks (SNNs) inherently convey information via dynamic spikes distributed over time~\cite{maass1997networks}, akin to the functioning of brain networks. They possess greater potential for simulating brain-inspired topologies and functions owing to their intricate spatial-temporal dynamics. Delving further into SNNs could potentially unveil insights into the enigmatic process of efficient decision-making within the brain~\cite{pfeiffer2018deep}.

\subsection{Integrating SNNs with DRL}

Recently, a growing body of literature has explored the integration of SNNs with DRL~\cite{DBLP:conf/ijcai/ZhangZJW022,wang2023complex,zhang2022tuning}. Several methods convert a trained DQN into a SNN version~\cite{patel2019improved,tan2021strategy} or directly train a deep spiking Q-network~\cite{liu2021human,chen2022deep}.
These endeavors yield competitive performance on Atari video games with discrete action spaces compared to the original DQN.
Another set of approaches presents a hybrid learning paradigm tested on robot control tasks with continuous action spaces~\cite{tang2020reinforcement,tang2021deep,zhang2021population,DBLP:conf/aaai/ZhangZJ022}. This paradigm entails training a SAN in combination with artificial critic networks utilizing policy-based DRL algorithms.

Nevertheless, both AANs and regular SANs are confined to inter-layer connections, simplifying dendritic trees as linear structures and lacking intra-layer connections, which somewhat restricts the network's expressivity. In contrast, the BPT-SAN models the local nonlinearity of dendritic trees in the inter-layer connections and introduces lateral interactions between adjacent neurons in the intra-layer connections. These two network topologies collaboratively enhance the network's information processing capacity, resulting in improved decision-making performance.

\begin{figure*}[t]
\centering
\includegraphics[width=1.0\textwidth]{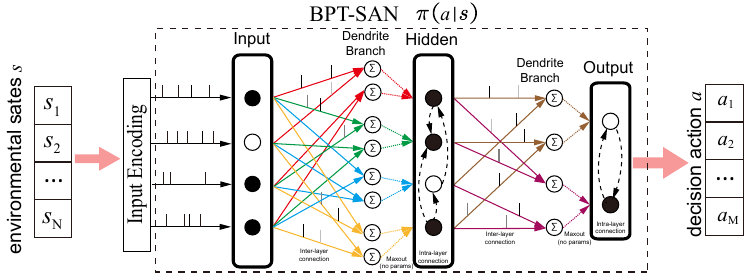} 
\caption{
The schematic diagram of our proposed BPT-SAN, which integrates spiking neurons with rich spatial-temporal dynamics and network topologies featuring biologically-plausible connectivity patterns. In the inter-layer connections, the BPT-SAN models the local nonlinearity of dendritic trees by breaking down the standard layer into two stages. In the initial stage, dendritic branches perform a mutually exclusive partition of the input and subsequently execute a weighted summation of the sparsely connected inputs. In the subsequent stage, the outputs of all branches converge to produce the neuron output via a maxout strategy. Furthermore, within the intra-layer connections, the BPT-SAN introduces lateral interactions to incorporate spiking states from neighboring neurons effectively. These two network topologies work synergistically to significantly enhance the information processing capacity of the network, enabling efficient decision-making in DRL.}
\label{fig-BPT-SAN}
\end{figure*}

\section{Method}

In RL tasks, an agent engages with the environment by selecting actions based on states and receiving corresponding rewards. 
At each time step $t$, the agent chooses action $\bm{a}^t$ from a policy $\pi$ contingent on the present state $\bm{s}^t$. 
Subsequently, the agent obtains a scalar reward $r^{t+1}$ and a new state $\bm{s}^{t+1}$, collectively forming an episode $\epsilon={\bm{s}^0,\bm{a}^0,r^1,...,\bm{s}^{T-1},\bm{a}^{T-1},r^T}$, where $T$ represents the episode's length. 
The return of episode $\epsilon$ is calculated as the sum of rewards: $R^{\epsilon}=\sum_{i=1}^Tr^t$, with the expectation that it will be larger for better outcomes.
 To establish a mapping from the state space to the action one, we construct a BPT-SAN that represents the policy $\pi$, predicting decision actions $\bm{a}^t$ based on environmental state $\bm{s}^t$, as depicted in Figure~\ref{fig-BPT-SAN}.

Before being fed into the BPT-SAN, the state undergoes encoding into a spike train through an input encoding module, aligning it with the inherent spatial-temporal dynamics of spiking neurons. 
The BPT-SAN features inter-layer connections enabled by nonlinear dendritic trees, as well as intra-layer connections facilitated by lateral interactions between neighboring neurons, synergistically enhancing the network's information processing capabilities. 
Ultimately, the output action is derived by decoding the average firing rate of accumulated spikes in a predefined time window.

\subsection{Input Encoding}
Consider a continuous state $\bm{s}\in \mathbb{R}^N$, encompassing details like joint angles, velocities, and external forces within a robotic control scenario.\footnote{For clarity, the time step $t$ of the RL episode is omitted.} This state is encoded into discrete spike trains $\bm{I}_\tau, \tau={1,2,..., T_1}$, where $T_1$ denotes the duration of the time window for the BPT-SAN.

Specifically, drawing inspiration from prior research~\cite{tang2021deep}, we initially translate the state $\bm{s}$ into stimulation strength $\bm{A}_{P}$ through population coding. Subsequently, the computed $\bm{A}_{P}$ is harnessed to generate the spike train $\bm{I}_\tau$ utilizing either Poisson coding or deterministic coding.

\paragraph{Population coding} We establish a neuron population $P_{i}$ dedicated to encoding each dimension of the state $\bm{s}_{i}$. Within this population, every neuron $P_{i,j}$ possesses a Gaussian receptive field featuring two adaptable parameters: mean $\mu_{i,j}$ and standard deviation $\sigma_{i,j}$. The process of population coding can be expressed as follows:
\begin{equation}
\left\{\begin{array}{l}
{A}_{P_{i,j}}=\exp^{-\frac{({s}_i-\mu_{i,j})^2}{2\sigma_{i,j}^2}} \\
\bm{A}_{P} = \left [{A}_{P_{1,1}},\dots,{A}_{P_{i,j}},\dots,{A}_{P_{N,J}} \right ]\\
\end{array}\right.
\end{equation}
where $i=1,...,N$ signifies the index for dimensions within the input state, $j=1,...,J$ denotes the neuron index within a population, and $\bm{A}_{P}\in\mathbb{R}^{N\cdot J}$ corresponds to the stimulation strength achieved through population coding.

\paragraph{Poisson coding} 
Based on the understanding that the Poisson process can be conceptualized as the culmination of a Bernoulli process, the probability-laden stimulation strength $\bm{A}_{P}$ can be employed to generate the binary random number. The ${I}_{\tau,k}$ will draw a value of $1$ based on the $k^{th}$ probability value ${A}_{P_{k}}$ derived from $\bm{A}_{P}$, formulated as:
\begin{equation}
P({I}_{\tau,k}=1) = C_R^r {A}_{P_{k}}^r\left(1-{A}_{P_{k}}\right)^{R-r} \\
\end{equation}
where $k=1,...,N$$\cdot$$J$ denotes the index for dimensions within the spike train $\bm{I}_\tau$.

\paragraph{Deterministic coding}

The stimulation strength $\bm{A}_{P}$ serves as the pre-synaptic inputs to the post-synaptic neurons, formulated as:
\begin{equation}
{v}'_{\tau,k} = {v}'_{\tau-1,k} + {A}_{P_{k}}
\end{equation}

\begin{equation}
{I}_{\tau,k} =
\left\{\begin{array}{l}
\begin{matrix}
     1  & \text{If}\ \ ({v}'_{\tau,k}>1)\\
     0  & \text{Else}
\end{matrix}
\end{array}\right.
\end{equation}
where ${v}'_{\tau,k}$ is reset as ${v}'_{\tau,k} - 1$ when ${I}_{\tau,k}=1$, ${v}'_{\tau,k}$ is pseudo membrane potential.

\subsection{BPT-SAN}
The BPT-SAN integrates spiking neurons with rich spatial-temporal dynamics and network topologies featuring biologically-plausible connectivity patterns.
It employs the current-based Leaky-Integrate-and-Fire (LIF) neuron~\cite{tang2021deep} as the foundation, which stands as one of the most commonly utilized artificial spiking neurons.
The dynamics of conventional LIF neurons~\cite{dayan2005theoretical} are typically orchestrated in two main steps: 1) integrating the incoming pre-synaptic spikes through inter-layer connections to generate current; and 2) processing this current to determine membrane potentials. In contrast, our approach involves an additional facet, wherein lateral interactions between adjacent neurons are also incorporated into membrane potentials via intra-layer connections.
Following this, the neuron generates a spike when its membrane potential surpasses a certain threshold. Here, we adopted the hard-reset mechanism, which involves resetting the membrane potential to its resting state once a spike is fired. These resulting spikes are then conveyed to the post-synaptic neurons during the same inference timestep.

For a specific LIF neuron $j$ at layer $l$, the entire process can be formulated as follows:
\begin{equation}
    \begin{aligned}
        c^{(l)}_{\tau,j} &= d_c\cdot c^{(l)}_{\tau-1,j} + \textbf{Inter}(\bm{O}^{(l-1)}_{\tau}) \\
        v^{(l)}_{\tau,j} &= d_v\cdot v^{(l)}_{\tau-1,j}\cdot(1-o^{(l)}_{\tau-1,j}) +  c^{(l)}_{\tau,j} +\textbf{Intra}(\bm{\mathcal{N}}^{(l)}_{j,\tau-1}) \\   
        o^{(l)}_{\tau,j} &= \mathbb{I}(v^{(l)}_{\tau,j} > v_{\text{th}})
    \end{aligned}
    \label{LIF}
\end{equation}
Here, the variables are defined as follows: $c^{(l)}_{\tau,j}$ signifies the current of neuron $j$ in layer $l$ at time $\tau$, with $d_c$ representing the current decay factor. $\textbf{Inter}(\bm{O}^{(l-1)}_{\tau})$ denotes the inter-layer computation utilizing the spiking state vector of all neurons in layer $l-1$ at time $\tau$ as input, while $\bm{O}^{(0)}_{\tau}=\bm{I}_\tau$. $v^{(l)}_{\tau,j}$ indicates the membrane potential of neuron $j$ in layer $l$ at time $\tau$, with $d_v$ representing the potential decay factor. $\textbf{Intra}(\bm{\mathcal{N}}^{(l)}_{j,\tau-1})$ encompasses the intra-layer computation employing the spiking state vector of neuron $j$'s neighboring neurons in layer $l$ at time $\tau-1$ as input. $o^{(l)}_{\tau,j}$ denotes the spike of neuron $j$ in layer $l$ at time $\tau$. $\mathbb{I}$ functions as the indicator function which means that when the input is \textit{True}, the output is $1$, and when the input is \textit{False}, the output is $0$. $v_{\text{th}}$ stands as the predefined firing threshold.

\paragraph{Inter-layer connections} Conventional AANs and regular SANs~\cite{tang2021deep} straightforwardly transmit weighted pre-synaptic spikes to post-synaptic neurons, simplifying dendritic trees to rudimentary linear structures. The inter-layer computation can be defined as:
\begin{equation}
    \textbf{Inter}(\bm{O}^{(l-1)}_{\tau}) = \bm{w}^{(l)}_j\bm{O}^{(l-1)}_{\tau}
    \label{full-connection}
\end{equation}
where $\bm{w}^{(l)}_j\in\mathbb{R}^{n_{l-1}}$ is the synaptic weight vector corresponding to neuron $j$ in layer $l$, with $n_{l-1}$ denoting the number of neurons in layer $l-1$.

Nonetheless, insights gleaned from both experimental and modeling investigations~\cite{losonczy2008compartmentalized,stuart2016dendrites} unveil that dendritic arbors house a profusion of active ion channels, rendering them superlinear and far more intricate than the simplistic weighted sum model outlined above.

Drawing inspiration from this insight, we model the local nonlinearity inherent in dendritic trees within the inter-layer connections. Depicted in Figure~\ref{fig-BPT-SAN}, each post-synaptic neuron $j=1,...,n_l$ within layer $l$ where $n_l$ signifies the neuron count in layer $l$, possesses $d$ dendritic branches.
Each branch establishes connections with $k$ inputs originating from the preceding layer $l-1$. 
The selection strategy entails each branch randomly opting for $k = n_{l-1}/d$ connections, drawn without replacement from the pool of $n_{l-1}$ inputs. 
This approach guarantees that each input spike associates with a dendrite branch just once, evading redundancy, and that the synapse sets of each branch remain mutually exclusive for every neuron. 
This selection method finds support in physiological studies suggesting that axons tend to refrain from forging multiple synaptic connections with the dendritic arbor of each pyramidal neuron~\cite{yuste2011dendritic,chklovskii2004cortical}. Additionally, \textbf{the mutually exclusive connection selection strategy underpins efficient network inference, as it keeps the same number of connections (parameters) as Equation (\ref{full-connection})}.

Formally, we denote $d$ as the branch number and $k$ as the branch size, adhering to the condition $k<n_{l-1}$. Pre-synaptic inputs first undergo a weighted summation at each branch, which is then followed by a maxout strategy to mold the ultimate output for the post-synaptic neuron. At this time, the inter-layer computation can be expressed as:
\begin{equation}
\begin{aligned}
\textbf{Inter}(\bm{O}^{(l-1)}_{\tau}) &= \text{maxout}\Big[g^{(l)}_{j,1}(\bm{O}^{(l-1)}_{\tau,1}),...,g^{(l)}_{j,d}(\bm{O}^{(l-1)}_{\tau,d})\Big]\\
g^{(l)}_{j,m}(\bm{O}^{(l-1)}_{\tau,m}) &= (\bm{\mathcal{M}}^{(l)}_{j,m}*\bm{W}^{(l)}_{j,m})(\bm{O}^{(l-1)}_{\tau,m}) 
\end{aligned}
\end{equation}
Here, $g^{(l)}_{j,m}(\bm{O}^{(l-1)}_{\tau,m})$ signifies the weighted summation on dendritic branch $m$ of neuron $j$ within layer $l$, utilizing the selected pre-synaptic input for each dendrite $\bm{O}^{(l-1)}_{\tau,m}\in\mathbb{R}^{k}$ as input, with $m=1,...,d$.
$\bm{W}^{(l)}_j\in\mathbb{R}^{n_{l-1}\times d}$ represents the weight matrix connecting pre-synaptic inputs to dendritic branches of post-synaptic neuron $j$ in layer $l$, while $\bm{\mathcal{M}}^{(l)}_j\in\mathbb{R}^{n_{l-1}\times d}$ corresponds to the accompanying mask matrix indicating whether a branch is linked to an input. The values within $\bm{\mathcal{M}}^{(l)}_j$ are binary ($1$ or $0$).
The matrix $\bm{\mathcal{M}}^{(l)}_j$ is constructed using a deterministic pseudo-random number generator, initialized with a predetermined random seed. Consequently, it can be reproduced directly from the original random seed, ensuring that generating $\bm{\mathcal{M}}^{(l)}_j$ doesn't impose additional model storage or transfer expenses, given a suitable algorithm.

\paragraph{Intra-layer connections} 
In contrast to conventional AANs and standard SANs~\cite{tang2021deep}, we introduce intra-layer connections that incorporate lateral interactions among neighboring neurons into the membrane potential to achieve more precise spike firing. This is expressed as follows:
\begin{equation}
   \textbf{Intra}(\bm{\mathcal{N}}^{(l)}_{j,\tau-1}) = \bm{w}'^{(l)}_j \bm{\mathcal{N}}^{(l)}_{j,\tau-1}
\end{equation}
where $\bm{w}'^{(l)}_j$ is the weight vector to integrate the spike states of neuron $j$'s neighboring neurons in layer $l$.

The BPT-SAN comprises multiple hidden layers, each featuring inter-layer localized nonlinearities and intra-layer lateral interactions that synergistically amplify the network's information processing capability.
Ultimately, spikes in the output layer are aggregated within a predetermined time window, enabling the computation of an average firing rate. This rate is subsequently harnessed for decoding a continuous action $\bm{a}\in\mathbb{R}^{M}$ containing the torque exerted on each joint of a robot.

\subsection{The Hybrid Learning of BPT-SAN}

The BPT-SAN demonstrates functional equivalence to both the AAN and regular SAN, rendering it compatible with any actor-critic-based policy gradient algorithm. To facilitate effective learning, we train the BPT-SAN in conjunction with artificial critic networks, employing the TD3~\cite{fujimoto2018addressing} and SAC~\cite{haarnoja2018soft} algorithms, respectively, within the recently introduced hybrid learning framework~\cite{tang2020reinforcement,tang2021deep}.
The TD3 algorithm employs twin Q-networks and delayed policy updates to enhance training stability and performance, whereas the SAC algorithm, a maximum entropy DRL approach, optimizes policies with heightened flexibility and stability by maximizing policy entropy. They are two state-of-the-art DRL algorithms for robotic control with continuous action space.

During hybrid training, our BPT-SAN deduces a output action $\bm{a}$ from an input state $\bm{s}$ to represent the policy, while artificial critic networks estimate the corresponding action-value $\mathit{Q}(\bm{s},\bm{a})$, guiding the BPT-SAN to acquire an improved policy. Following training, we assess the trained BPT-SAN across various robot control benchmarks, comparing its performance against its AAN counterpart and regular SAN within the same experimental settings. For a detailed training procedure of the TD3 and SAC algorithms, please refer to~\cite{fujimoto2018addressing} and~\cite{haarnoja2018soft}, respectively.

\subsection{Tuning BPT-SAN with Pseudo Backpropagation}

Given that the spike firing process in Equation~(\ref{LIF}) lacks differentiability, direct tuning of the BPT-SAN using the Backpropagation (BP) algorithm~\cite{rumelhart1986learning} is unfeasible. This challenge is addressed by employing the pseudo BP approach~\cite{zenke2018superspike,wang2023attention}, which involves substituting the non-differentiable components of spiking neurons during BP with a pre-defined gradient value.
Here, we adopt the equation of the rectangular function to approximate the gradient of a discrete spike, outlined as follows:
\begin{equation}
    z(v) = \begin{cases}
        1\ \ & \text{if}\ |v-v_{\text{th}}| < w\\
        0\ \ & \text{otherwise}
    \end{cases}
\end{equation}
where $z$ denotes a pseudo-gradient, $v$ represents the membrane potential, $v_{\text{th}}$ is the predefined firing threshold, and $w$ signifies the threshold window for gradient passage.

\begin{figure}[tbp]
	\centering  
	\includegraphics[width=0.86\columnwidth]{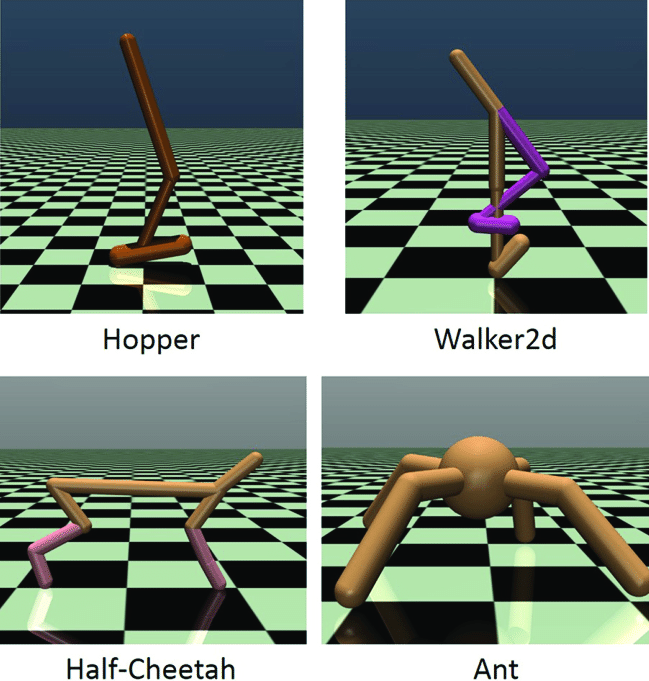}
 	\caption{Four continuous control tasks. \textbf{(a) Hopper}: State dimension: $N=11$, Action dimension: $M=3$, Goal: make a 2D one-legged robot hop forward as fast as possible; \textbf{(b) Walker2d}: State dimension: $N=17$, Action dimension: $M=6$, Goal: make a 2D bipedal robot walk forward as fast as possible; \textbf{(c) Half-Cheetah}: State dimension: $N=17$, Action dimension: $M=6$, Goal: make a 2D cheetah robot run as fast as possible; \textbf{(d) Ant}: State dimension: $N=111$, Action dimension: $M=8$, Goal: make a four-legged creature walk forward as fast as possible.}
  \label{fig-task}
\end{figure}

\section{Experimental Settings}

\subsection{Evaluation Tasks}
We evaluate the performance of our BPT-SAN on a suit of continuous control tasks interfaced by OpenAI Gym~\cite{brockman2016openai}, as depicted in Figure~\ref{fig-task}. These tasks encompass Hopper-v3, Walker2d-v3, Half-Cheetah-v3, and Ant-v3, all executed within the fast physics simulator MuJoCo~\cite{todorov2012mujoco}.

\begin{figure*}[t]
	\centering
	\includegraphics[width=1.0\textwidth]{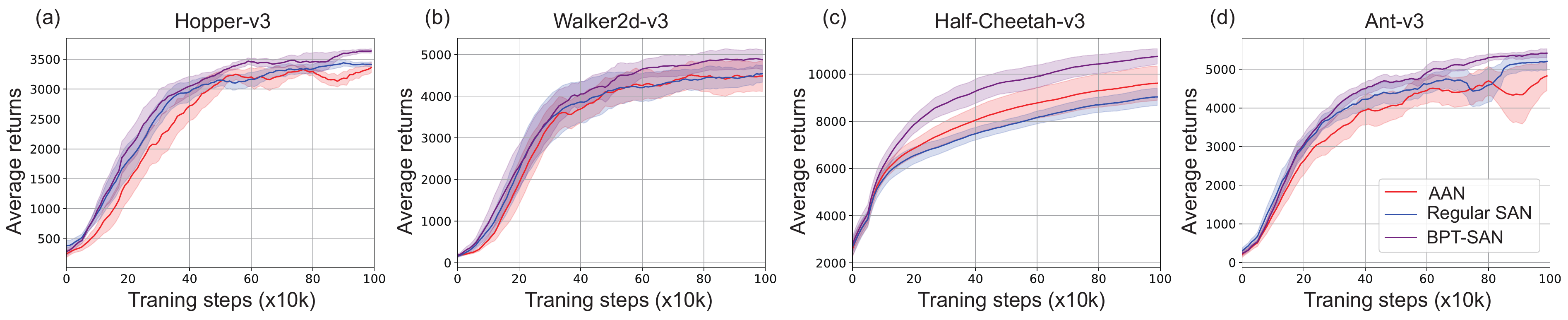}
	\caption{Comparison of average returns achieved by various actor networks trained with \textbf{the TD3 algorithm}. (a) Performance of AAN, Regular SAN, and BPT-SAN during training on the Hopper-v3 task. (b, c, d) Performances of these three actor networks on Walker2d-v3, Half-Cheetah-v3, and Ant-v3, respectively. Notably, our BPT-SAN outperforms the others consistently across all tasks. The shaded area illustrates half a standard deviation of the average evaluation result across $10$ random seeds, while the curves are smoothed for improved visualization.
 }
 \label{fig-main-td3}
\end{figure*}

\begin{figure*}[t]
	\centering
	\includegraphics[width=1.0\textwidth]{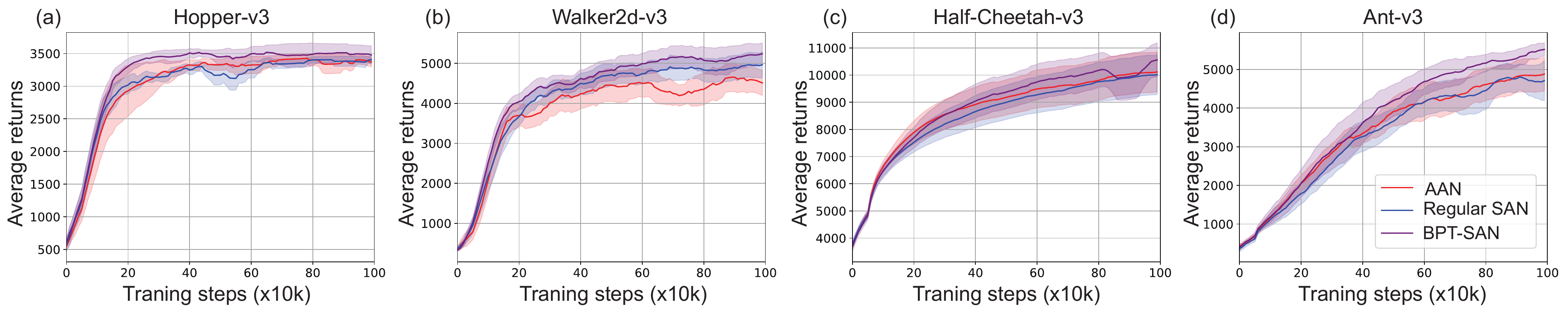}
	\caption{Comparison of average returns achieved by various actor networks trained with \textbf{the SAC algorithm}. (a) Performance of AAN, Regular SAN, and BPT-SAN during training on the Hopper-v3 task. (b, c, d) Performances of these three actor networks on Walker2d-v3, Half-Cheetah-v3, and Ant-v3, respectively. Notably, our BPT-SAN outperforms the others consistently across all tasks. The shaded area illustrates half a standard deviation of the average evaluation result across $10$ random seeds, while the curves are smoothed for improved visualization.}\label{fig-main-sac}
\end{figure*}

\subsection{Implement Details}

We compare our BPT-SAN against the conventional AAN and regular SAN\footnote{Referred to as PopSAN in the original paper~\cite{tang2021deep}.}.
AAN, regular SAN, and our BPT-SAN are all trained in combination with artificial critic networks of the same structure (\emph{i.e.}, two $256$-unit hidden layers with ReLU activation function and one $1$-unit output layer) using the TD3 and SAC algorithms, respectively.
The implementation and hyper-parameter configuration of TD3 and SAC are from the open-source OpenAI Spinning Up\footnote{\url{https://github.com/openai/spinningup}}.
We evaluate the trained AAN, regular SAN, and our BPT-SAN on the above four robot control tasks within the identical configurations and compare their achieved returns.
Other hyper-parameter configurations are as follows:

The AAN consists of two hidden layers and one output layer, with each hidden layer comprising $256$ artificial neurons using the ReLU activation function~\cite{glorot2011deep}. It has only inter-layer connections, implemented through direct linear weighted summation, and lacks intra-layer connections.
The regular SAN and BPT-SAN share the same number of layers and neurons with the AAN, but utilize the current-based LIF neuron as their foundation.
For the LIF neuron, the duration of the time window $T_1$ is $5$, the current decay factor $d_c$ is $0.5$, the potential decay factor $d_v$ is $0.75$, the resting potential is $0$, the firing threshold $v_\text{th}$ is $0.5$, and the threshold window for passing the gradient $w$ is $0.5$.
The regular SAN employs a linear weighted summation structure for inter-layer connections, lacking intra-layer connections. In contrast, our BPT-SAN utilizes a nonlinear dendritic tree structure with a branch number $d=2$ for inter-layer connections and introduces intra-layer connections to integrate spiking information from neighboring neurons. Both regular SAN and BPT-SAN initially utilize population encoding to encode states, with the number of neurons per population, $J$, set to $10$. For the Hopper-v3 and Walker2d-v3 tasks, spike trains are generated using Poisson coding, while deterministic coding is used for the Ant-v3 and Half-Cheetah-v3 tasks. Finally, actions are decoded based on the average spike firing rate within the time window.

To address concerns regarding reproducibility~\cite{henderson2018deep}, all experiments are reported over $10$ random seeds of the gym simulator and network initialization. 
Each task undergoes $1$ million steps of execution and is assessed at intervals of $10$k steps.
Each evaluation computes the average return across $10$ episodes, with each episode spanning a maximum of $1000$ execution steps.
All experiments are conducted on an AMD EPYC $7742$ server equipped with $3$ NVIDIA A100 GPUs, each with $40$GB of memory.

\begin{figure*}[t]
	\centering
	\includegraphics[width=1.0\textwidth]{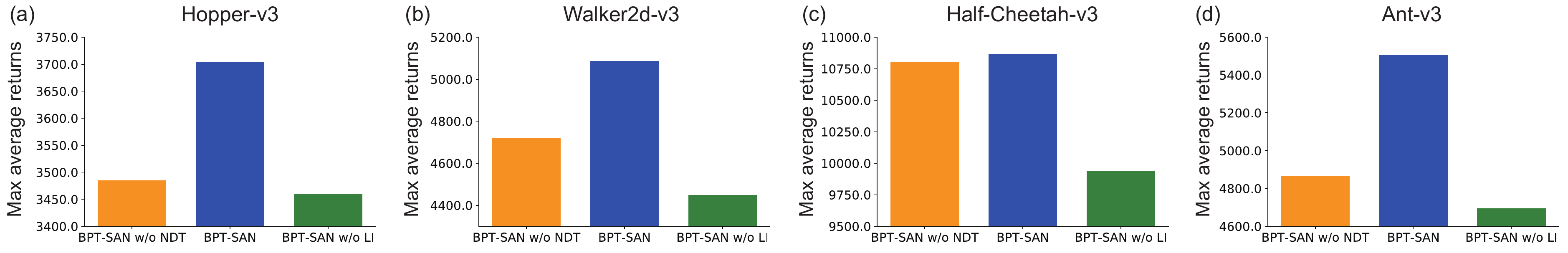}
	\caption{Comparison of the max average returns over $10$ random seeds achieved by our BPT-SAN and its two variants: BPT-SAN w/o NDT (replace Nonlinear Dendritic Tree (NDT) with linear weighted summation in inter-layer connections) and BPT-SAN w/o LI (remove Lateral Interaction (LI) in the intra-layer connections), all trained using \textbf{the TD3 algorithm}. (a) Performance of BPT-SAN w/o NDT, BPT-SAN, and BPT-SAN w/o LI on the Hopper-v3 task. (b, c, d) Performances of these three actor networks on Walker2d-v3, Half-Cheetah-v3, and Ant-v3, respectively. It's worth noting that the BPT-SAN, featuring both inter-layer NDT and intra-layer LI, consistently outperforms the other two variants across all tasks, highlighting the superiority of this comprehensive architecture.
}
 \label{fig-ab-td3}
\end{figure*}

\begin{figure*}[t]
	\centering
	\includegraphics[width=1.0\textwidth]{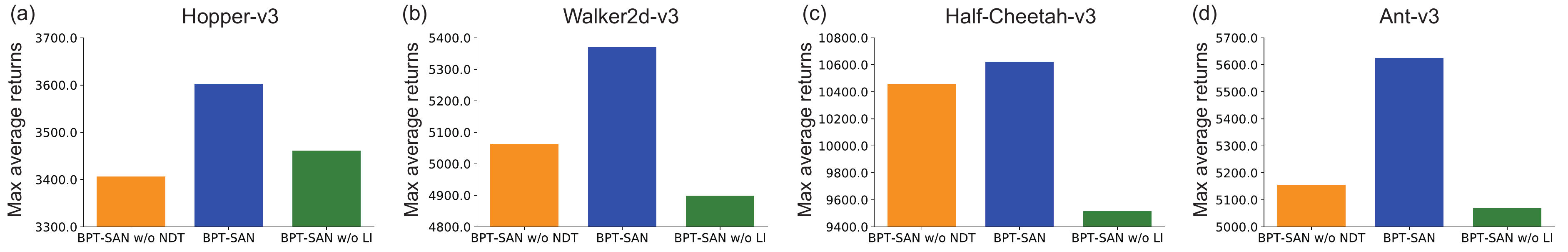}
	\caption{Comparison of the max average returns over $10$ random seeds achieved by our BPT-SAN and its two variants: BPT-SAN w/o NDT and BPT-SAN w/o LI, all trained using \textbf{the SAC algorithm}. (a) Performance of BPT-SAN w/o NDT, BPT-SAN, and BPT-SAN w/o LI on Hopper-v3. (b, c, d) Performances of these three actor networks on Walker2d-v3, Half-Cheetah-v3, and Ant-v3, respectively. It's worth noting that the BPT-SAN, featuring both inter-layer NDT and intra-layer LI, consistently outperforms the other two variants across all tasks, highlighting the superiority of this comprehensive architecture.}\label{fig-ab-sac}
\end{figure*}

\section{Results and Discussions}

\subsection{Main Results}

We conduct a comprehensive performance comparison between our BPT-SAN, trained with both the TD3 and SAC algorithms, and its AAN counterpart and regular SAN.
As shown in Figures~\ref{fig-main-td3} and~\ref{fig-main-sac}, our BPT-SAN consistently outperforms the other actor networks on all tasks, showcasing the efficacy of the proposed BPT-SAN for continuous robot control scenarios.
Remarkably, our AAN and SAN baselines achieve comparable or superior performance to what has been reported in the existing literature~\cite{fujimoto2018addressing,haarnoja2018soft,tang2021deep}.

\subsection{Ablation Study}

We conduct a thorough comparison between our BPT-SAN and its two distinct variants: BPT-SAN w/o NDT and BPT-SAN w/o LI, trained using the TD3 and SAC algorithms, respectively. Specifically, BPT-SAN w/o NDT involves the elimination of the Nonlinear Dendritic Tree (NDT) structure from the inter-layer connection, replacing it with a direct linear weighted summation. Similarly, BPT-SAN w/o LI removes the Lateral Interaction (LI) from the intra-layer connection of BPT-SAN.
Illustrated in Figures~\ref{fig-ab-td3} and~\ref{fig-ab-sac}, our BPT-SAN, equipped with both inter-layer NDT and intra-layer LI, consistently surpasses the other two variants across all tasks. This consistent superiority indicates that these two network topologies synergistically enhance the network's expressivity and information processing capacity.

\section{Conclusion}

In this paper, drawing inspiration from brain networks, we introduce the BPT-SAN, an innovative framework that seamlessly integrates spiking neurons with complex spatial-temporal dynamics and network topologies featuring biologically-plausible connectivity patterns for efficient decision-making in DRL.
The BPT-SAN models the intricate nonlinearity of dendritic trees in inter-layer connections and incorporates lateral interactions among neighboring neurons in intra-layer connections. These two network topologies collaboratively enhance the network's expressivity and information processing capacity, achieving a better performance across four continuous control tasks from OpenAI Gym than its AAN counterpart and regular SAN.

In the future, a captivating and promising direction lies in the deeper integration of biologically-plausible principles inspired by brain networks into the SAN. By embracing these principles, the horizon opens up to the possibility of achieving not only enhanced performance but also more energy-efficient computations, heightened adaptability to dynamic environments, and heightened robustness in decision-making. The interplay between neuroscience and artificial intelligence holds great potential for shaping the future landscape of research and innovation.

\newpage

\bibliography{aaai24}

\begin{thebibliography}{50}
\providecommand{\natexlab}[1]{#1}

\bibitem[{Bellemare et~al.(2013)Bellemare, Naddaf, Veness, and
  Bowling}]{bellemare2013arcade}
Bellemare, M.~G.; Naddaf, Y.; Veness, J.; and Bowling, M. 2013.
\newblock {The arcade learning environment: An evaluation platform for general
  agents}.
\newblock \emph{Journal of Artificial Intelligence Research}, 47: 253--279.

\bibitem[{Brockman et~al.(2016)Brockman, Cheung, Pettersson, Schneider,
  Schulman, Tang, and Zaremba}]{brockman2016openai}
Brockman, G.; Cheung, V.; Pettersson, L.; Schneider, J.; Schulman, J.; Tang,
  J.; and Zaremba, W. 2016.
\newblock {Openai gym}.
\newblock \emph{arXiv preprint arXiv:1606.01540}.

\bibitem[{Chen et~al.(2022)Chen, Peng, Huang, and Tian}]{chen2022deep}
Chen, D.; Peng, P.; Huang, T.; and Tian, Y. 2022.
\newblock {Deep Reinforcement Learning with Spiking Q-learning}.
\newblock \emph{arXiv preprint arXiv:2201.09754}.

\bibitem[{Cheng et~al.(2020)Cheng, Hao, Xu, and Xu}]{cheng2020lisnn}
Cheng, X.; Hao, Y.; Xu, J.; and Xu, B. 2020.
\newblock {LISNN: Improving spiking neural networks with lateral interactions
  for robust object recognition.}
\newblock In \emph{IJCAI}, 1519--1525. Yokohama.

\bibitem[{Dayan and Abbott(2005)}]{dayan2005theoretical}
Dayan, P.; and Abbott, L.~F. 2005.
\newblock \emph{{Theoretical neuroscience: computational and mathematical
  modeling of neural systems}}.
\newblock MIT press.

\bibitem[{Douglas and Martin(2004)}]{douglas2004neuronal}
Douglas, R.~J.; and Martin, K.~A. 2004.
\newblock {Neuronal circuits of the neocortex}.
\newblock \emph{Annu. Rev. Neurosci.}, 27: 419--451.

\bibitem[{Duan et~al.(2016)Duan, Chen, Houthooft, Schulman, and
  Abbeel}]{duan2016benchmarking}
Duan, Y.; Chen, X.; Houthooft, R.; Schulman, J.; and Abbeel, P. 2016.
\newblock {Benchmarking deep reinforcement learning for continuous control}.
\newblock In \emph{International conference on machine learning}, 1329--1338.
  PMLR.

\bibitem[{Fujimoto, Hoof, and Meger(2018)}]{fujimoto2018addressing}
Fujimoto, S.; Hoof, H.; and Meger, D. 2018.
\newblock {Addressing function approximation error in actor-critic methods}.
\newblock In \emph{International Conference on Machine Learning}, 1587--1596.
  PMLR.

\bibitem[{Glorot, Bordes, and Bengio(2011)}]{glorot2011deep}
Glorot, X.; Bordes, A.; and Bengio, Y. 2011.
\newblock {Deep sparse rectifier neural networks}.
\newblock In \emph{Proceedings of the fourteenth international conference on
  artificial intelligence and statistics}, 315--323. JMLR Workshop and
  Conference Proceedings.

\bibitem[{Haarnoja et~al.(2018)Haarnoja, Zhou, Abbeel, and
  Levine}]{haarnoja2018soft}
Haarnoja, T.; Zhou, A.; Abbeel, P.; and Levine, S. 2018.
\newblock {Soft actor-critic: Off-policy maximum entropy deep reinforcement
  learning with a stochastic actor}.
\newblock In \emph{International conference on machine learning}, 1861--1870.
  PMLR.

\bibitem[{Henderson et~al.(2018)Henderson, Islam, Bachman, Pineau, Precup, and
  Meger}]{henderson2018deep}
Henderson, P.; Islam, R.; Bachman, P.; Pineau, J.; Precup, D.; and Meger, D.
  2018.
\newblock {Deep reinforcement learning that matters}.
\newblock In \emph{Proceedings of the AAAI conference on artificial
  intelligence}, volume~32.

\bibitem[{Jiang, Xu, and Liang(2017)}]{jiang2017deep}
Jiang, Z.; Xu, D.; and Liang, J. 2017.
\newblock {A deep reinforcement learning framework for the financial portfolio
  management problem}.
\newblock \emph{arXiv preprint arXiv:1706.10059}.

\bibitem[{Kaelbling, Littman, and Moore(1996)}]{kaelbling1996reinforcement}
Kaelbling, L.~P.; Littman, M.~L.; and Moore, A.~W. 1996.
\newblock {Reinforcement learning: A survey}.
\newblock \emph{Journal of artificial intelligence research}, 4: 237--285.

\bibitem[{Koch, Poggio, and Torre(1983)}]{koch1983nonlinear}
Koch, C.; Poggio, T.; and Torre, V. 1983.
\newblock {Nonlinear interactions in a dendritic tree: localization, timing,
  and role in information processing.}
\newblock \emph{Proceedings of the National Academy of Sciences}, 80(9):
  2799--2802.

\bibitem[{Lillicrap et~al.(2016)Lillicrap, Hunt, Pritzel, Heess, Erez, Tassa,
  Silver, and Wierstra}]{lillicrap2016continuous}
Lillicrap, T.~P.; Hunt, J.~J.; Pritzel, A.; Heess, N.; Erez, T.; Tassa, Y.;
  Silver, D.; and Wierstra, D. 2016.
\newblock {Continuous control with deep reinforcement learning.}
\newblock In \emph{ICLR (Poster)}.

\bibitem[{Liu et~al.(2021)Liu, Deng, Xie, Huang, and Tang}]{liu2021human}
Liu, G.; Deng, W.; Xie, X.; Huang, L.; and Tang, H. 2021.
\newblock {Human-Level Control through Directly-Trained Deep Spiking
  Q-Networks}.
\newblock \emph{arXiv preprint arXiv:2201.07211}.

\bibitem[{Losonczy, Makara, and Magee(2008)}]{losonczy2008compartmentalized}
Losonczy, A.; Makara, J.~K.; and Magee, J.~C. 2008.
\newblock {Compartmentalized dendritic plasticity and input feature storage in
  neurons}.
\newblock \emph{Nature}, 452(7186): 436--441.

\bibitem[{Maass(1997)}]{maass1997networks}
Maass, W. 1997.
\newblock {Networks of spiking neurons: the third generation of neural network
  models}.
\newblock \emph{Neural networks}, 10(9): 1659--1671.

\bibitem[{Mel and Svoboda(2004)}]{chklovskii2004cortical}
Mel, B.; and Svoboda, K. 2004.
\newblock {Cortical rewiring and information storage}.
\newblock \emph{Nature}, 431(7010): 782--788.

\bibitem[{Mnih et~al.(2016)Mnih, Badia, Mirza, Graves, Lillicrap, Harley,
  Silver, and Kavukcuoglu}]{mnih2016asynchronous}
Mnih, V.; Badia, A.~P.; Mirza, M.; Graves, A.; Lillicrap, T.; Harley, T.;
  Silver, D.; and Kavukcuoglu, K. 2016.
\newblock {Asynchronous methods for deep reinforcement learning}.
\newblock In \emph{International conference on machine learning}, 1928--1937.
  PMLR.

\bibitem[{Mnih et~al.(2015)Mnih, Kavukcuoglu, Silver, Rusu, Veness, Bellemare,
  Graves, Riedmiller, Fidjeland, Ostrovski et~al.}]{mnih2015human}
Mnih, V.; Kavukcuoglu, K.; Silver, D.; Rusu, A.~A.; Veness, J.; Bellemare,
  M.~G.; Graves, A.; Riedmiller, M.; Fidjeland, A.~K.; Ostrovski, G.; et~al.
  2015.
\newblock {Human-level control through deep reinforcement learning}.
\newblock \emph{nature}, 518(7540): 529--533.

\bibitem[{Patel et~al.(2019)Patel, Hazan, Saunders, Siegelmann, and
  Kozma}]{patel2019improved}
Patel, D.; Hazan, H.; Saunders, D.~J.; Siegelmann, H.~T.; and Kozma, R. 2019.
\newblock {Improved robustness of reinforcement learning policies upon
  conversion to spiking neuronal network platforms applied to Atari Breakout
  game}.
\newblock \emph{Neural Networks}, 120: 108--115.

\bibitem[{Pfeiffer and Pfeil(2018)}]{pfeiffer2018deep}
Pfeiffer, M.; and Pfeil, T. 2018.
\newblock {Deep learning with spiking neurons: Opportunities and challenges}.
\newblock \emph{Frontiers in neuroscience}, 12: 774.

\bibitem[{Ratliff, Hartline, and Lange(1966)}]{ratliff1966dynamics}
Ratliff, F.; Hartline, H.~K.; and Lange, D. 1966.
\newblock {The dynamics of lateral inhibition in the compound eye of Limulus.
  I}.
\newblock \emph{The functional organization of the compound eye}, 399--424.

\bibitem[{Rumelhart, Hinton, and Williams(1986)}]{rumelhart1986learning}
Rumelhart, D.~E.; Hinton, G.~E.; and Williams, R.~J. 1986.
\newblock {Learning representations by back-propagating errors}.
\newblock \emph{nature}, 323(6088): 533--536.

\bibitem[{Schulman et~al.(2015)Schulman, Levine, Abbeel, Jordan, and
  Moritz}]{schulman2015trust}
Schulman, J.; Levine, S.; Abbeel, P.; Jordan, M.; and Moritz, P. 2015.
\newblock {Trust region policy optimization}.
\newblock In \emph{International conference on machine learning}, 1889--1897.
  PMLR.

\bibitem[{Schulman et~al.(2017)Schulman, Wolski, Dhariwal, Radford, and
  Klimov}]{schulman2017proximal}
Schulman, J.; Wolski, F.; Dhariwal, P.; Radford, A.; and Klimov, O. 2017.
\newblock {Proximal policy optimization algorithms}.
\newblock \emph{arXiv preprint arXiv:1707.06347}.

\bibitem[{Schultz(1998)}]{schultz1998predictive}
Schultz, W. 1998.
\newblock {Predictive reward signal of dopamine neurons}.
\newblock \emph{Journal of neurophysiology}.

\bibitem[{Sehnke et~al.(2010)Sehnke, Osendorfer, R{\"u}ckstie{\ss}, Graves,
  Peters, and Schmidhuber}]{sehnke2010parameter}
Sehnke, F.; Osendorfer, C.; R{\"u}ckstie{\ss}, T.; Graves, A.; Peters, J.; and
  Schmidhuber, J. 2010.
\newblock {Parameter-exploring policy gradients}.
\newblock \emph{Neural Networks}, 23(4): 551--559.

\bibitem[{Silver et~al.(2016)Silver, Huang, Maddison, Guez, Sifre, Van
  Den~Driessche, Schrittwieser, Antonoglou, Panneershelvam, Lanctot
  et~al.}]{silver2016mastering}
Silver, D.; Huang, A.; Maddison, C.~J.; Guez, A.; Sifre, L.; Van Den~Driessche,
  G.; Schrittwieser, J.; Antonoglou, I.; Panneershelvam, V.; Lanctot, M.;
  et~al. 2016.
\newblock {Mastering the game of Go with deep neural networks and tree search}.
\newblock \emph{nature}, 529(7587): 484--489.

\bibitem[{Stuart, Spruston, and H{\"a}usser(2016)}]{stuart2016dendrites}
Stuart, G.; Spruston, N.; and H{\"a}usser, M. 2016.
\newblock \emph{{Dendrites}}.
\newblock Oxford University Press.

\bibitem[{Sutton and Barto(2018)}]{sutton2018reinforcement}
Sutton, R.~S.; and Barto, A.~G. 2018.
\newblock \emph{{Reinforcement learning: An introduction}}.
\newblock MIT press.

\bibitem[{Tan, Patel, and Kozma(2021)}]{tan2021strategy}
Tan, W.; Patel, D.; and Kozma, R. 2021.
\newblock {Strategy and Benchmark for Converting Deep Q-Networks to
  Event-Driven Spiking Neural Networks}.
\newblock In \emph{Proceedings of the AAAI Conference on Artificial
  Intelligence}, volume~35, 9816--9824.

\bibitem[{Tang, Kumar, and Michmizos(2020)}]{tang2020reinforcement}
Tang, G.; Kumar, N.; and Michmizos, K.~P. 2020.
\newblock {Reinforcement co-learning of deep and spiking neural networks for
  energy-efficient mapless navigation with neuromorphic hardware}.
\newblock In \emph{2020 IEEE/RSJ International Conference on Intelligent Robots
  and Systems (IROS)}, 6090--6097. IEEE.

\bibitem[{Tang et~al.(2021)Tang, Kumar, Yoo, and Michmizos}]{tang2021deep}
Tang, G.; Kumar, N.; Yoo, R.; and Michmizos, K. 2021.
\newblock {Deep Reinforcement Learning with Population-Coded Spiking Neural
  Network for Continuous Control}.
\newblock In \emph{Conference on Robot Learning}, 2016--2029. PMLR.

\bibitem[{Todorov, Erez, and Tassa(2012)}]{todorov2012mujoco}
Todorov, E.; Erez, T.; and Tassa, Y. 2012.
\newblock {Mujoco: A physics engine for model-based control}.
\newblock In \emph{2012 IEEE/RSJ international conference on intelligent robots
  and systems}, 5026--5033. IEEE.

\bibitem[{Van~Hasselt, Guez, and Silver(2016)}]{van2016deep}
Van~Hasselt, H.; Guez, A.; and Silver, D. 2016.
\newblock {Deep reinforcement learning with double q-learning}.
\newblock In \emph{Proceedings of the AAAI conference on artificial
  intelligence}, volume~30.

\bibitem[{Vinyals et~al.(2019)Vinyals, Babuschkin, Czarnecki, Mathieu, Dudzik,
  Chung, Choi, Powell, Ewalds, Georgiev et~al.}]{vinyals2019grandmaster}
Vinyals, O.; Babuschkin, I.; Czarnecki, W.~M.; Mathieu, M.; Dudzik, A.; Chung,
  J.; Choi, D.~H.; Powell, R.; Ewalds, T.; Georgiev, P.; et~al. 2019.
\newblock {Grandmaster level in StarCraft II using multi-agent reinforcement
  learning}.
\newblock \emph{Nature}, 575(7782): 350--354.

\bibitem[{Wang et~al.(2023{\natexlab{a}})Wang, Zhang, Zhang, and
  Xu}]{wang2023attention}
Wang, Q.; Zhang, D.; Zhang, T.; and Xu, B. 2023{\natexlab{a}}.
\newblock {Attention-free spikformer: Mixing spike sequences with simple linear
  transforms}.
\newblock \emph{arXiv preprint arXiv:2308.02557}.

\bibitem[{Wang et~al.(2023{\natexlab{b}})Wang, Zhang, Han, Wang, Zhang, and
  Xu}]{wang2023complex}
Wang, Q.; Zhang, T.; Han, M.; Wang, Y.; Zhang, D.; and Xu, B.
  2023{\natexlab{b}}.
\newblock {Complex dynamic neurons improved spiking transformer network for
  efficient automatic speech recognition}.
\newblock In \emph{Proceedings of the AAAI Conference on Artificial
  Intelligence}, volume~37, 102--109.

\bibitem[{Wang et~al.(2016)Wang, Schaul, Hessel, Hasselt, Lanctot, and
  Freitas}]{wang2016dueling}
Wang, Z.; Schaul, T.; Hessel, M.; Hasselt, H.; Lanctot, M.; and Freitas, N.
  2016.
\newblock {Dueling network architectures for deep reinforcement learning}.
\newblock In \emph{International conference on machine learning}, 1995--2003.
  PMLR.

\bibitem[{Watkins and Dayan(1992)}]{watkins1992q}
Watkins, C.~J.; and Dayan, P. 1992.
\newblock {Q-learning}.
\newblock \emph{Machine learning}, 8(3): 279--292.

\bibitem[{Wu et~al.(2018)Wu, Liu, Li, and Wu}]{wu2018improved}
Wu, X.; Liu, X.; Li, W.; and Wu, Q. 2018.
\newblock {Improved expressivity through dendritic neural networks}.
\newblock \emph{Advances in neural information processing systems}, 31.

\bibitem[{Yuste(2011)}]{yuste2011dendritic}
Yuste, R. 2011.
\newblock {Dendritic spines and distributed circuits}.
\newblock \emph{Neuron}, 71(5): 772--781.

\bibitem[{Zenke and Ganguli(2018)}]{zenke2018superspike}
Zenke, F.; and Ganguli, S. 2018.
\newblock {Superspike: Supervised learning in multilayer spiking neural
  networks}.
\newblock \emph{Neural computation}, 30(6): 1514--1541.

\bibitem[{Zhang et~al.(2022{\natexlab{a}})Zhang, Zhang, Jia, Wang, and
  Xu}]{DBLP:conf/ijcai/ZhangZJW022}
Zhang, D.; Zhang, T.; Jia, S.; Wang, Q.; and Xu, B. 2022{\natexlab{a}}.
\newblock {Recent Advances and New Frontiers in Spiking Neural Networks}.
\newblock In \emph{Proceedings of the Thirty-First International Joint
  Conference on Artificial Intelligence, {IJCAI} 2022, Vienna, Austria, 23-29
  July 2022}, 5670--5677.

\bibitem[{Zhang et~al.(2022{\natexlab{b}})Zhang, Zhang, Jia, Wang, and
  Xu}]{zhang2022tuning}
Zhang, D.; Zhang, T.; Jia, S.; Wang, Q.; and Xu, B. 2022{\natexlab{b}}.
\newblock {Tuning Synaptic Connections instead of Weights by Genetic Algorithm
  in Spiking Policy Network}.
\newblock \emph{arXiv preprint arXiv:2301.10292}.

\bibitem[{Zhang et~al.(2022{\natexlab{c}})Zhang, Zhang, Jia, and
  Xu}]{DBLP:conf/aaai/ZhangZJ022}
Zhang, D.; Zhang, T.; Jia, S.; and Xu, B. 2022{\natexlab{c}}.
\newblock {Multi-Sacle Dynamic Coding Improved Spiking Actor Network for
  Reinforcement Learning}.
\newblock In \emph{Thirty-Sixth {AAAI} Conference on Artificial Intelligence,
  {AAAI} 2022, Thirty-Fourth Conference on Innovative Applications of
  Artificial Intelligence, {IAAI} 2022, The Twelveth Symposium on Educational
  Advances in Artificial Intelligence, {EAAI} 2022 Virtual Event, February 22 -
  March 1, 2022}, 59--67. {AAAI} Press.

\bibitem[{Zhang, Zhang et~al.(2021)}]{zhang2021population}
Zhang, D.; Zhang, T.; et~al. 2021.
\newblock {Population-coding and Dynamic-neurons improved Spiking Actor Network
  for Reinforcement Learning}.
\newblock \emph{ArXiv preprint arXiv:2106.07854}.

\bibitem[{Zhao et~al.(2023)Zhao, Zhang, Han, Zhang, and Xu}]{zhao2023ode}
Zhao, X.; Zhang, D.; Han, L.; Zhang, T.; and Xu, B. 2023.
\newblock {ODE-based Recurrent Model-free Reinforcement Learning for POMDPs}.
\newblock \emph{Advances in Neural Information Processing Systems}, 25:
  27159--27170.

\end{thebibliography}

\end{document}